\documentclass{llncs}
\usepackage{makeidx}
\usepackage[utf8]{inputenc}
\usepackage{enumitem}
\usepackage{amsfonts}
\usepackage{amssymb}
\usepackage{amsmath}
\usepackage{graphicx}
\usepackage{subfigure}
\usepackage{color}
\usepackage{pifont}
\usepackage{multirow}
\usepackage{booktabs}
\begin{document}
\frontmatter
\pagestyle{headings}
\mainmatter
\title{Semi-supervised Learning for Word Sense Disambiguation
	\thanks{This work was awarded the Third Place in the EST 2013 Contest (ISSN 1850-2946) at the 42nd JAIIO (Annals of 42nd JAIIO - Argentine Journals of Informatics - ISSN 1850-2776).  
The original article in Spanish, titled ``Desambiguaci\'{o}n de Palabras Polis\'{e}micas mediante Aprendizaje Semi-supervisado'', can be found here: \url{http://42jaiio.sadio.org.ar/proceedings/simposios/Trabajos/EST/20.pdf}.
	}
}
\author{Dar\'{i}o Garigliotti}
\institute{FAMAF - Faculty of Mathematics, Astronomy, Physics and Computation\\
National University of Córdoba, Argentina\\
\email{dario.g.professional@gmail.com}
}

\maketitle

\begin{abstract}

This work is a study of the impact of multiple aspects in a classic unsupervised word sense disambiguation algorithm.
We identify relevant factors in a decision rule algorithm, including the initial labeling of examples, the formalization of the rule confidence, and the criteria for accepting a decision rule.  
Some of these factors are only implicitly considered in the original literature.  
We then propose a lightly supervised version of the algorithm, and employ a pseudo-word-based strategy to evaluate the impact of these factors.  
The obtained performances are comparable with those of highly optimized formulations of the word sense disambiguation method.
\end{abstract}

%
\section{Introduction} \label{section-intro}

Word sense disambiguation consists in determining the correct sense of a particular occurrence of a polysemous word.  
Disambiguating word occurrences is a key task for many basic Natural Language Processing (NLP) components, since they assume that there is no ambiguity in an input text.  
A strategy widely adopted by the methods for this task is to emulate the disambiguation process carried out by humans when reading, this is, by using information about the context of a word occurrence.

This work is an evaluation of the impact of several aspects in a classic unsupervised word sense disambiguation algorithm.
By a close inspection of this algorithm, we identify factors that result relevant for its performance, some of which are only implicitly considered in the original literature where the algorithm is firstly described.  
We propose a lightly supervised version of the decision rule algorithm for word sense disambiguation.  
We make use of an evaluation strategy based on pseudo-words to show that our approach achieves comparable performance to those of highly optimized versions of the algorithm.

The rest of this paper is organized as follows.  
Section~\ref{section-relevant} describes related literature in word sense disambiguation, in particular, the unsupervised algorithm studied in this work.  
Our semi-supervised decision rule approach is then presented in Section~\ref{section-methods}.  
Next, Section~\ref{section-results} discusses the experimental results.  
Finally, in Section~\ref{section-conclusions} we present the conclusions, and mention a number of directions for future investigation.  

%
\section{Related Work} \label{section-relevant}
Word sense disambiguation has been addressed with supervised learning methods.  
These require a large dataset of examples, typically sentences with the occurrence of an ambiguous word, labeled by hand with the correct sense.

Yarowsky~\cite{yar} introduces an unsupervised algorithm that combines the structure of decision lists with the unsupervised labeling of a large amount of initial examples.  
Given an ambiguous word, referred to as target, this labeling is performed by determining a seed collocation or evidence that is representative of each of the senses of the target.  
In this way, it preserves the lack of supervision, yet it is not fully unsupervised since it requires labeled examples.  
Moreover, it is possible to introduce bias in the choice of the initial collocations.  
This method, hereinafter referred to as the Yarowsky algorithm, uses two main heuristics: (i) \emph{one sense per collocation}, meaning that words co-occurring with the target give strong clues about its correct sense, and (ii) \emph{one sense per discourse}, by which it can be assumed that the target has the same sense in multiple occurrences within the same discourse or text.  
The work by Abney~\cite{abn} is the first systematic study of this algorithm.  

%
\section{Approach} \label{section-methods}
The main part of this section details the unsupervised algorithm introduced by Yarowsky, where we also discuss the identified factors and the approach we obtain.  
This section ends with the description of the processing steps used to obtain the dataset from a textual corpus.  

\subsection{Word Sense Disambiguation Algorithm} \label{subsection:architecture-algor}
Given a set of instances, each with an occurrence of a fixed target word, the Yarowsky algorithm consists in performing the following steps:

\begin{itemize}
  \item[$\bullet$] Determine representative collocations, and split the instances set in subsets according to the sense labels.
  \item[$\bullet$] While the non-labeled instances set does not converge:
  \begin{itemize}
    \item[-] Learn a decision rule of the shape \textit{collocation} $\Rightarrow$  \textit{sense}, by inspecting the labeled examples.
    \item[-] Add to the decision list the rules that meet the acceptance criteria.
    \item[-] Sort the decision list by rule confidence.
    \item[-] Apply the decision list to the entire instances set, labeling each example with the first rule in the list that meet the rule condition.
    \item[-] Apply the ``one sense per discourse'' criterion to possibly filter out examples labeled in the step above.
  \end{itemize}
  \item[$\bullet$] After converging to a residual set of non-labeled examples, the final rule decision list can be applied on examples not seen yet.
\end{itemize}

In the algorithm introduced by Yarowsky~\cite{yar}, many criteria are slightly justified, and if so, appear without their respective (optimal) parameters.  
Other important algorithm factors are not even mentioned explicitly.  
Specifically, in this work we identify the following factors, and propose the respective experimental settings for our version of the algorithm:

\begin{itemize}[leftmargin=*]
  \item[$\bullet$] Factor 1: Unlike relying on seed collocations as previously mentioned, we perform the \emph{initial labeling} by manually disambiguating two examples per sense, for each target word.  In this way, the learning algorithm becomes semi-supervised.  As we show in Section~\ref{section-results}, this setting is very important.  

  \item[$\bullet$] Factor 2: Regarding \textit{types of collocations or evidences}, Yarowsky considers co-occurrence adjacent to any other word in the context (typically a sentence surrounding the target), co-occurrence adjacent to content words (i.e., nouns, adjectives, verbs or adverbs) versus function words, and co-occurrence in a window of +/- $n$ to $m$ words, for given $n, m$.  The original algorithm analysis does not mention the impact of which of those collocation types to use, nor with which parameters, e.g., for $n, m$ in the windows-based case.  In this work, we only use simple co-occurrence collocations, i.e., a word co-occurring with the target in any position in the context.  

  \item[$\bullet$] Factor 3: We discard the ``one sense per discourse'' heuristics.
  
  \item[$\bullet$] Factor 4: The \emph{equation of rule confidence}, or rule probability, that we consider is different from the classic based on log-likelihood that is used in the original algorithm~\cite{yar}.  Given $f(E_i)$, the number of examples with the collocation or evidence $E_i$, and $f(S_j,E_i)$, the number of examples with evidence $E_i$ labeled with the sense $S_j$, an optimization~\cite{tsu:chi} allows us to use the following equation of the confidence that a given evidence determines a given sense:
\end{itemize}

  \newcommand\restr[2]{{
    \left.\kern-\nulldelimiterspace
    #1
    \vphantom{\big|}
    \right|_{#2}
  }}
  
\begingroup
  \setlength{\abovedisplayskip}{0pt plus 1pt minus 1pt}
  \begin{equation} \label{eq:reliability-1}
    \text{confidence} = \frac{f(S_j,E_i)}{f(S_j,E_i) + f(\neg S_j,E_i)} = \frac{f(S_j,E_i)}{\left.f\right|_{\{labeled \; examples\}}(E_i)}~.
  \end{equation}
\endgroup

\setlength{\belowdisplayskip}{10.0pt plus 2.0pt minus 5.0pt}

\begin{itemize}
  \item[$\bullet$] Factor 5: The Yarowsky algorithm does not mention the confidence \textit{threshold} as part of the acceptance criteria for a decision rule into the decision list.  Abney~\cite{abn} proposes a threshold equal to 1/$L$, with $L$ being the number of senses known for the target.  We work with a very strict confidence threshold initially equal to 0.95.  
  
  \item[$\bullet$] Factor 6: Meanwhile Yarowsky~\cite{yar} allows to remove the label of an example when the rule that gave it such a label falls below the threshold, Abney~\cite{abn} forces to always preserve a label once an example has been disambiguated, but the label can be changed.  In our work, a labeled example remains with its first assigned sense, i.e., it cannot be removed or changed.  
  
  \item[$\bullet$] Factor 7: In the case of having the same confidence, we sort two rules according to a second sorting criteria, the one of \textit{coverage}, this is, the number of examples whose collocation meets the rule condition.  Note that rule coverage is equal to the denominator in Eq.~\eqref{eq:reliability-1}.  This factor is not mentioned in the Yarowsky algorithm~\cite{yar}, yet it is considered by Tsuruoka and Chikayama~\cite{tsu:chi}.
\end{itemize}
\subsection{Dataset Construction}  \label{dataset-construction}

Given a textual corpus, we process it to construct the dataset as follows.

\begin{itemize}
	\item[$\bullet$] We retain every line of raw text in the corpus only if it is made of at least ten words.
	\item[$\bullet$] We manually disambiguate two sentences per sense for each given target word, and label accordingly in the corpus.  We refer sometimes to this stage as the initial training.  
	\item[$\bullet$] We perform lemmatization and part-of-speech (POS) tagging.  Since we worked with a corpus in Spanish, in this step we apply the FreeLing\footnote{\url{http://nlp.lsi.upc.edu/freeling/index.php/node/1}} 2.2.2 tool.
	\item[$\bullet$] We preserve only content words, i.e., nouns, main verbs, and qualificative adjectives.  These correspond, respective, with the POS-tags of the shape N****, VM****, and AQ****, in the parole-EAGLES standard.\footnote{\url{https://www.cs.upc.edu/\%7Enlp/tools/parole-sp.html}}  
	\item[$\bullet$] We split this lemmatized, POS-tagged corpus into sentences, i.e., contexts.
	\item[$\bullet$] We select those contexts where the target occurs.
	\item[$\bullet$] We identify the contexts manually disambiguated in the first step.
	\item[$\bullet$] We take the set of every different lemma occurring in any of these contexts.
	\item[$\bullet$] We build our lexicon, by restricting the set of lemmas defined above to contain a lemma only if it occurs in at least ten contexts.
	\item[$\bullet$] We sort the lexicon according to the number of contexts each lemma occurs in.
	\item[$\bullet$] We vectorize the corpus according to the lexicon truncated by frequency of its lemmas, to obtain a training dataset where each instance has a counter of co-occurrences of each lemma with the target in that context.  
\end{itemize}

Two assumptions are worthy to be mentioned regarding the steps described above.  
First, by performing lemmatization, it is intuitively assumed that morphological accidents do not alter the sense of the word in that context.  
Second, the lexicon is truncated as a usual practice due to the sparseness phenomenon.  This happens when the vocabulary is very large and the words occurring in a context are generally different from the ones used in another context.  

%
\section{Experimental Results} \label{section-results}
In this section, we describe our evaluation strategy, and the experimental settings, as well as we analyze the experimental results.  

\subsection{Evaluation} \label{subsection:architecture-eval}
The evaluation strategy based on pseudo-words was introduced by Schütze~\cite{sch} as a simple and very economic method to evaluate word sense disambiguation algorithms.  
This method consists in choosing randomly two words, for example ``banana'' and ``door,'' and replacing in a text corpus every occurrence of any of the two by the new target pseudo-word ``bananadoor.''  
A word sense disambiguation algorithm is applied on the selected contexts.  
After this, each example is considered correctly disambiguated if the assigned sense (``banana'' or ``door'') coincides with the original word that was replaced in that context by the pseudo-word.  
Even though the sense ambiguity introduced by this method can be seen as artifactual, the advantage is to produce large amounts of labeled examples for evaluation with almost no cost.  
This evaluation method also shows the independence of the disambiguation algorithm with respect to the language of the corpus which is applied on, since it only uses the impact of the given collocations and does not assume any convention or bias in the language.  
Given the two words that are often chosen to explain the pseudo-word replacement, this evaluation method in general is also known as \emph{bananadoor evaluation}.  

We employ two simple word sense disambiguation algorithms with which to compare the performance of our proposed approach:

\begin{itemize}
	\item[$\bullet$] Baseline: given the word most often replaced by the pseudo-word ---e.g., ``door''---, corresponding to a $k$\% of all the replacements, the baseline labels every context with the most frequent sense ---in this example, ``door''--- and obtains an accuracy of $k$\%.  
	\item[$\bullet$] Random: using the same information that $s$ is the most frequent sense with $k$\% of the replacements, this algorithm labels each context probabilistically, assigning the sense $s$ a $k$\% of the times.  
\end{itemize}

\subsection{Experimental Settings} \label{subsection:environment}
We work with a corpus of 57 million words in Spanish language, consisting of digital articles of the Spanish newspapers La Vanguardia and El Periódico de Catalunya.  
For simplicity, our problem is constrained to disambiguating the sense of words with the following properties:

\begin{itemize}
  \item[$\bullet$] Every target has only two senses, and are generally very different.
  \item[$\bullet$] Every target is a noun, and all its senses are nouns.  
  \item[$\bullet$] Possibly, targets considered polysemous include cases of words that are actually homonyms.
\end{itemize}

In a first part of our experiments, we obtain a set of contexts for each target, in a set of five selected targets.  
We refer to the contexts set of each target as the target dataset.  
We observe the impact of the identified factors in the disambiguation performance, in particular, analyzing, for example, the number of iterations required for convergence.  
Table~\ref{table:targets-senses-sizes} presents statistics of the datasets for the five selected targets.  
The two senses of each of the binary targets observed in this part are referred to as senses A and B.  

\begin{table}[t]
	\caption{The five binary targets and statistics of their datasets.}
	\centering
	\begin{tabular}{ l @{\hskip 0.3in}l @{\hskip 0.3in}l @{\hskip 0.2in}r @{\hskip 0.2in}r }
		\toprule
		Target & Sense A & Sense B & Dataset size & Lexicon size \\
		\midrule
		Manzana & Fruta & Superficie & 712 & 144 \\
		Naturaleza & Índole & Entorno & 2,607 & 611 \\
		Movimiento & Cambio & Corriente & 6,509 & 1,883 \\
		Tierra & Materia & Planeta & 7,874 & 2,019 \\
		Interés & Finanzas & Curiosidad & 14,640 & 3,104 \\
		\bottomrule
	\end{tabular}
	\label{table:targets-senses-sizes}
\end{table}

In a second part, we obtain a new dataset, by applying the bananadoor evaluation.  
The replacements are made using two words in Spanish relatively frequent, ``vida'' (Spanish for ``life'') and ``ciudad'' (``city''), leading to the pseudo-word ``vidaciudad.''
This datasets consists of 62,819 examples and a lexicon of 3,937 lemmas.  
Here, the size of the lexicon after truncation is considerably smaller than the sizes for the datasets per target in the first part, which may result of high importance when applying clustering on the dataset without enough computational resources.  

\subsection{Analysis} \label{subsection:analysis}
In the first part, we apply our approach to each target dataset.  
The results in Table~\ref{table:targets-convergences-ratios} show a quick convergence into a residual, stable set of non-labeled examples.  
Clearly, any factor which increases the number of rules in the decision list will have a positive impact regarding this convergence.  
For example, a possible one would be the Factor 1: to perform the initial labeling with the original algorithm leads to more rules for the first iteration.  
Another way would involve Factor 3: by implementing also the ``one sense per discourse'' criterion, each iteration possibly provides more labeled examples to the next iteration.  
In both cases, for a larger dataset, it may lead to less iterations, although each one would require longer since it would have to inspect more rules.  
Then, these factors, as assumed in our experimental setting, have a negative impact in the final residual set.  
Factor 2 is of positive impact regarding convergence, due to the fact that a larger number of collocations are able to capture more reliably some linguistic phenomena that escape from the simple co-occurrence setting.  

\begin{table}[t]
	\caption{Convergence, and proportions of final residual sets of examples.}
	\centering
	\begin{tabular}{p{3.cm} c c c c c c }
		\toprule
		& \multicolumn{6}{c}{Target} \\
		& manzana & naturaleza & movimiento & tierra & interés & vidaciudad \\
		\midrule
		No. of iterations needed to converge & 5 & 6 & 7 & 8 & 7 & 6 \\
		\midrule
		\% of the final residual set w.r.t. the dataset & 6.46\% & 33.10\% & 42.83\% & 18.97\% & 33.25\% & 77.62\% \\
		\bottomrule
	\end{tabular}
	\label{table:targets-convergences-ratios}
\end{table}

Our experimental setting for Factor 6 does not allow re-labeling.  
This may suggest a positive impact, yet re-labeling is taken under consideration in the previous work.  Then, it should be closely inspected regarding a sort of ``speed versus accuracy'' dilemma.  
In Factor 5, a rule confidence threshold too lenient can impact positively in the convergence yet negatively in the accuracy.  
An element already mentioned yet not explicitly identified among our factors is the lexicon truncation bound, presented in the last column of Table~\ref{table:targets-senses-sizes}.  
This bound affects the performance in the second part, when trying to disambiguate the pseudo-target ``vidaciudad.'' Specifically, it leaves a large final residual set since we require that the lemmas of its lexicon occur each in at least 30 contexts.  
Lowering the bound improves this situation, but going down excessively in reducing this bound allows for entrance of noise from unfrequent lemmas.

\begin{figure}[t]
	\centering
	\subfigure[Proportion of decision rules.]{\includegraphics[width=0.48\linewidth]{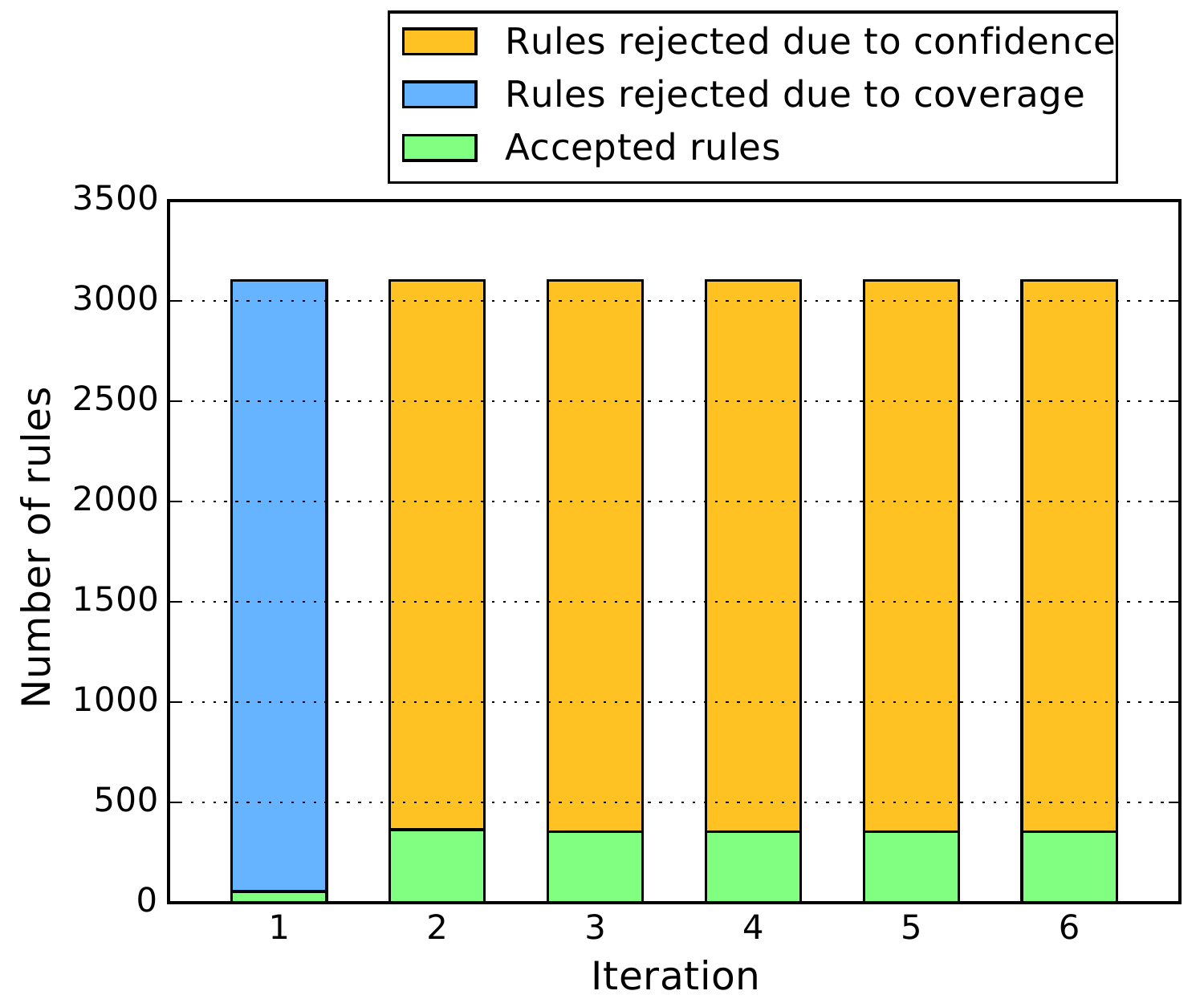} \label{fig:histograms-1}}
	\hspace{0.03in}
	\subfigure[Subsets of examples.]{\includegraphics[width=0.48\linewidth]{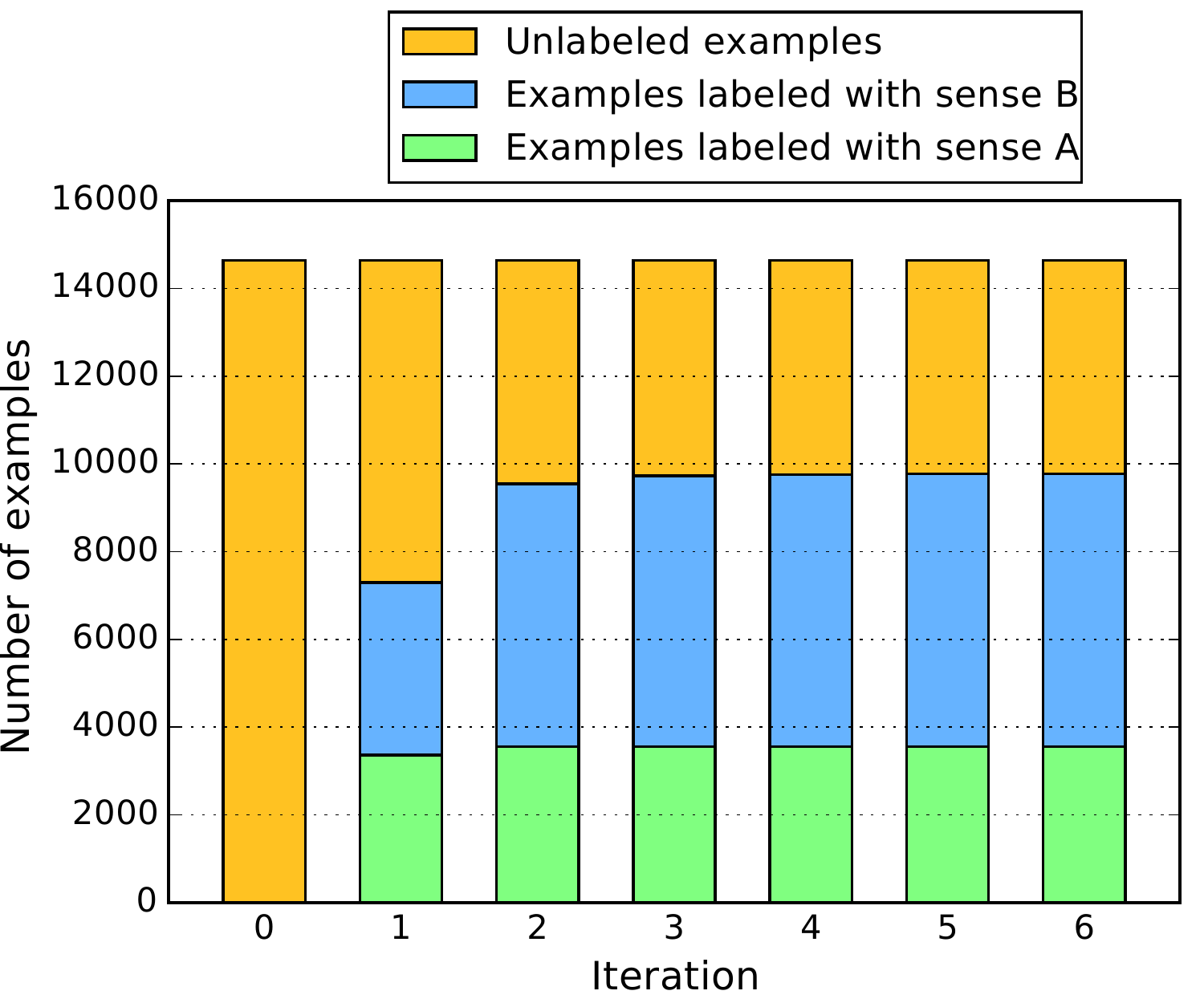} \label{fig:histograms-2}}
	\caption{Performance of our approach per iteration, for the target word ``interés.''}
	\label{fig:histograms}
\end{figure}

As pointed out in the related work~\cite{tsu:chi}, Eq.~\eqref{eq:reliability-1} can lead to undesired situations, given the few evidences available at the beginning of the learning problem.     
Figure~\ref{fig:histograms-1} shows the performance per iteration, regarding the proportions of decision rules.  
We can see that the first iteration of the algorithm accepts very few rules, just within the required coverage, while most of the rules are rejected.  
The coverage criterion used in our experimental settings (Factor 7) requires as few as at least one evidence; experiments with any stricter setting results harmful due to the sparseness phenomenon, as it rejects every rule in the first two iterations and converges to a final set without any newly labeled examples.  

In summary, performance is very sensitive to Factor 7, since any stricter setting would lead to no disambiguation altogether.  
This factor is in close relation with our setting of very light supervision for the initial training set.  
Factor 1, i.e., semi-supervised learning, is crucial in our approach.  
As we can observe in Fig.~\ref{fig:histograms-2}, the initial rule decision list is drastically changed in the first iteration w.r.t. the proportion of labeled examples in the dataset, and after that, the example subsets become stable until convergence as the rules get refined due to a larger coverage.

\begin{table*}[t]
	\caption{Performance of our approach (Decision List), and clustering performance (average across the five targets).}
	\centering
	\begin{tabular}{p{2.cm}|p{1.6cm}p{1.6cm}p{2.3cm}|p{1.6cm}p{2.3cm}}
		\toprule
		& \multicolumn{3}{|c|}{Bananadoor Evaluation} & \multicolumn{2}{|c}{Clustering (average)} \\
		\midrule
		Algorithm & Baseline & Random & Decision List & KMeans & EM (2 clusters) \\
		\midrule
		Accuracy & 51.10\% & 50.13\% & \textbf{59.86\%} & 72.9534\% & 72.4977\% \\
		\bottomrule
	\end{tabular}
	\label{table:performances-eval-cluster}
\end{table*}

Factor 4, the confidence equation, has a particular impact in relationship with the Factor 7 of coverage.  
A formalization, given by Tsuruoka and Chikayama~\cite{tsu:chi} ---see Eq.~\eqref{eq:reliability-5} in Appendix~\ref{appendix-equation}--- calculates the optimal smoothing for Eq.~\eqref{eq:reliability-1}; this optimization is obtained to overcome the problem of low initial coverage.  
When using the optimized formulation, most of the collocations in the first iteration have zero coverage, and by smoothing, they receive a portion of the large confidence that the very few rules with non-zero coverage have.  
Yet, since the rules with null coverage are so many, in this way each of these rules gets a probability less than $10^{-3}$, i.e., they are rejected by our threshold, as they will be by any other reasonable confidence threshold.  

Moving to the second part of our experiments, we conduct the bananadoor evaluation.
Table~\ref{table:performances-eval-cluster} presents the experimental results in terms of accuracy.  
Our decision list approach outperforms the baseline and random algorithms.  
Furthermore, we can say that it is a reasonable performance when comparing with the accuracy of 69.4\%, achieved by a variant of the literature~\cite{tsu:chi} that employs a log-likelihood-based confidence formulation and optimized coverage of at least three evidences.
As shown in the last columns of Table~\ref{table:performances-eval-cluster}, the accuracy is still higher when observing the average clustering accuracy of the five targets from the first part of our experiments.  
Nevertheless, it is worthy to mention that, due to limitations in the computational resources, clustering was applied on a restricted version of the datasets, made of only the ten most frequent lemmas of the respective lexicons.  
We would expect a performance significantly lower in the case of clustering the entire dataset, due to the noise introduced by using the full lexicon.

%
\section{Conclusions} \label{section-conclusions}
We have conducted an evaluation of factors relevant to the performance of a lightly supervised word sense disambiguation algorithm.  
Our experimental results indicate that the initial training is a crucial element, regarding both the convergence and accuracy.  
It is indeed so relevant, that it affects the space of parameters for other factors of large impact, such as the confidence threshold and coverage.  
We also observe that an optimization of the confidence equation with smoothing can result harmful to the performance, due to the same consequences of the initial labeling.  
Other factors like re-labeling, ``one sense per discourse'' criterion, or involving more collocation types, may improve the performance, at the cost of a slower convergence.  

There are several additional lines of study that we are interested to explore in future work.  
Firstly, we could observe in a closer look the impact of some factors identified in this work.  For example, to consider a variant of our approach that also includes rules according to adjacency of a lemma that belongs to a particular morphosyntactic category.  An instance of this would be the qualificative adjective ``human'' as a fixed lemma adjacent to ``nature'' in a phrase like ``human nature,'' which results in a very distinctive lemma for the ``condition'' sense of ``nature.''
Secondly, to introduce a criterion for the coverage factor that makes its setting dynamic, so that it becomes stricter as long as the iterations progress.  In this way, the population of rules could be controlled, by restricting them to only the more reliable ones.
Another line of future investigation is to observe the results using an initial training coupled with a misleading bias, this is, picking contexts containing collocations close to the target that suggest the sense different from the one manually labeled.  
This bias ---and the converse one, i.e., picking sentences very representative of a sense for the target--- is likely of high impact, yet with our strategy of initial training, it results a challenge to avoid every possible bias.  
Lastly, another line of future work is to perform, previous to the application of the word sense disambiguation algorithm, a stage of sense discovery.  
This could be done, for example, by clustering the original, unlabeled dataset, leading to a more natural label set partition.  
This line of investigation is of particular interest since it involves one of the key problems in word sense disambiguation, this is, defining the domain of senses.  

%

%
\newpage

%
\appendix

%
\section{Optimization of the Confidence Rule (Tsuruoka y Chikayama~\cite{tsu:chi})} \label{appendix-equation}
The equation used in the original description of the Yarowsky algorithm~\cite{yar} to calculate the confidence of a decision rule is:

\begin{equation} \label{eq:reliability-2}
  \text{confidence} = \log \left(\frac{P(S_j | E_i)}{P(\neg S_j | E_i)}\right)~,
\end{equation}

where $E_i$ is the contextual evidence or collocation, and $S_j$ the candidate sense for labeling an example.  
Tsuruoka y Chikayama~\cite{tsu:chi} obtain another formulation, that produce decision lists equivalent to the ones from Eq.~\eqref{eq:reliability-2}:  

\begin{equation} \label{eq:reliability-3}
  \text{confidence} = P(S_j | E_i)~.
\end{equation}

When a large number of examples with evidence $E_i$ are available,  Eq.~\eqref{eq:reliability-3} can be estimated, using Maximum Likelihood, as follows:

\begin{equation} \label{eq:reliability-4}
  P(S_j | E_i) = \frac{f(S_j,E_i)}{f(E_i)}~,
\end{equation}

where $f(E_i)$ is the number of examples with evidence $E_i$, and $f(S_j,E_i)$ is the number of examples with evidence $E_i$ and labeled with sense $S_j$.  

The equation that we use in this work is Eq.~\ref{eq:reliability-1}:

\begin{equation}
  \text{confidence} = \frac{f(S_j,E_i)}{f(S_j,E_i) + f(\neg S_j,E_i)} = \frac{f(S_j,E_i)}{\left.f\right|_{\{labeled \; examples\}}(E_i)}~. \nonumber
\end{equation}

This formulation overestimates Eq.~\eqref{eq:reliability-4} by restricting the denominator to only the set of labeled examples.  
Equation~\ref{eq:reliability-1} is slightly corrected in the actual implementation of the algorithm to avoid division by zero.  

Using Bayesian learning, Tsuruoka and Chikayama~\cite{tsu:chi} obtain a formulation of the beta distribution that involves $\Theta = P(S_j | E_i)$, and so they estimate the optimal \textit{smoothing}:

\begin{equation} \label{eq:reliability-5}
  \text{confidence} = E[\Theta]= \frac{f(S_j,E_i) + 1}{f(E_i) + 2}~.
\end{equation}
\\
\\
\\
\\
\\
\\
\\
\centering
\ding{91} \ding{91} \ding{91}

\begin{thebibliography}{}
%
\bibitem{yar}
David Yarowsky.  
1995.  
Unsupervised word sense disambiguation rivaling supervised methods.
In \emph{Proc. of the 33rd Annual Meeting of the Association for Computational Linguistics (ACL).}  
189--196.  

\bibitem{abn}
Steven Abney.  
2004.  
Understanding the Yarowsky Algorithm.  
\emph{Computational Linguistics, 30 (3).}  
365--395.  

\bibitem{tsu:chi}
Yoshimasa Tsuruoka and Takashi Chikayama.
2001.  
Estimating Reliability of Contextual Evidences in Decision-List Classifiers under Bayesian Learning.  
In \emph{Proc. of the Sixth Natural Language Processing Pacific Rim Symposium (NLPRS).}  
701--707.  

\bibitem{sch}
Hinrich Schütze.  
1992.
Context space.
\emph{AAAI Fall Symposium on Probabilistic Approaches to Natural Language.}  
113--120.  

\end{thebibliography}
\end{document}